\documentclass[10pt]{article} % For LaTeX2e

\usepackage[accepted]{rlj} % Should be uncommented for the camera-ready
\usepackage{amssymb}            % Defines common symbols like \mathbb R
\usepackage{mathtools}          % Extends amsmath, providing common math tools
\usepackage{mathrsfs}           % Enables \mathscr, which can work in cases that \mathcal does not
\usepackage{graphicx}           % For including images
\usepackage{subcaption}         % Allows for the use of subfigures and subcaptions
\usepackage[space]{grffile}     % For spaces in image names
\usepackage{url}                % For displaying URLs
\usepackage{lipsum}             % For placeholder text

\title{Analogy making as amortised model construction}%construal}
\setrunningtitle{RLC 2025 - Finding the Frame workshop}

\author{David G. Nagy\textsuperscript{1,2}, Tingke Shen\textsuperscript{2}, Hanqi Zhou\textsuperscript{1,2}, \mbox{Charley M. Wu\textsuperscript{1,2,3,4,\textdagger}},\\ \mbox{Peter Dayan\textsuperscript{2,1, \textdagger}}}

\emails{david.nagy@uni-tuebingen.de}

\affiliations{
$^{1}$\textbf{University of T{\"u}bingen, Germany}\\
$^{2}$\textbf{Max Planck Institute for Biological Cybernetics, Germany}\\
$^{3}$\textbf{Technical University of Darmstadt, Darmstadt, Germany}\\
$^{4}$\textbf{Hessian.AI, Darmstadt, Germany}
\par % If including additional comments like below, use \par to add some whitespace. 
$^\dagger$ Equal contribution
}

\begin{document}

%\makeCover  % Create the cover page
\maketitle  % Make the title section

\begin{abstract}
Humans flexibly construct internal models to navigate novel
situations. To be useful, these internal models must be sufficiently faithful to the environment that resource-limited planning leads to adequate outcomes; equally, they must be tractable to construct in the first place. We argue that analogy plays a central role in these processes, enabling agents to reuse solution-relevant structure from past experiences and amortise the computational costs of both model construction (construal) and planning. Formalising analogies as partial homomorphisms between Markov decision processes, we sketch a framework in which abstract modules, derived from previous construals, serve as composable building blocks for new ones. This modular reuse allows for flexible adaptation of policies and representations across domains with shared structural essence. 
\end{abstract}

%%%%%%%%%%%%%%%%%%%%%%%%%%%%%%%%%%%%%%%%%%%%%%%%%%%%%%%%%%%%%%%%
%% Section: Submission of papers to RLJ/RLC
%%%%%%%%%%%%%%%%%%%%%%%%%%%%%%%%%%%%%%%%%%%%%%%%%%%%%%%%%%%%%%%%
\section{Introduction}

Humans, like artificial reinforcement learning (RL) agents, maintain internal representations of their environment, enabling them to interpret sensory inputs and predict action outcomes. In RL, these representations usually take the form of Markov decision processes (MDPs\footnote{We use MDP here as a shorthand for the broader family of Markov decision process-based models, including extensions such as POMDPs (partially observable MDPs), BAMDPs (Bayes-adaptive MDPs), and IPOMDPs (interactive POMDPs), which allow for uncertainty, learning, and multi-agent reasoning, respectively.}) --- abstract formalisations of the environment that support planning and decision making. Typically, formulating such an MDP --- defining the states, observations, actions, rewards --- is  performed manually by an external designer \citep[although recent approaches allow the learning of this model from experience in restricted domains;][]{schrittwieser2020mastering,hafner2025mastering}.
For example, a vacuum-cleaning robot might perceive its environment in terms of ‘obstacles,’ ‘dirt concentration,’ and ‘returning to the charging dock’---rather than ‘minimal yet cozy living rooms,’ ‘ringing phones,’ or ‘following one’s dreams’---reflecting the designer’s assumptions about which abstractions are relevant for vacuuming floors. 

Unlike a robot vacuum confined to navigating an apartment, humans must operate in vastly more diverse situations: navigating not just an apartment, but an entire city (on foot, by bike, or by public transport); decorating an apartment or building a shelter, working in an office, negotiating, interpreting social nuances, or solving math problems. Crucially, the choice of the abstract representation used in these situations is deeply intertwined with decision-making processes operating over it --- an effective representation can render difficult problems trivially easy to solve, while a poor one can make a solution impossible \citep{giunchiglia_theory_1992, abel2022theory, ravindran_smdp_2003}.

We argue that the vast diversity of situations humans encounter rules out reliance on a pre-designed MDP --- or even a predetermined set of MDPs. Instead, humans must be capable of constructing MDPs for novel situations \textit{on-demand}. In other words, they must fill the shoes of not just the {user}, but also the {designer}, of their own internal MDPs --- yielding what we refer to throughout this paper as situation specific \textit{construals}, following \citet{ho_people_2022}. Both these roles have to be played in the face of limited cognitive resources. Unfortunately, the obvious solution of simplification leads to variants of the notoriously difficult ``frame problem'' \citep{dennett_cognitive_1990,icard_resource-rational_2015}.

Here, we take inspiration from a long body of work focusing on the central role of analogy and metaphor in human cognition \citep{hofstadter_surfaces_2013,lakoff_metaphors_2008,gentner_analogy_2017}. We propose that analogies (understood broadly) underlie the human ability for on-demand model construction, and more generally offer a powerful mechanism by which RL agents can flexibly adapt past knowledge to novel circumstances. Specifically, we formalise potential analogies as mappings between MDPs that preserve solution-relevant structure (and in some cases the entire solution). Through such analogical mappings, humans can construct new internal models by adapting and combining abstract MDPs that were effective in past situations. This reuse enables the transfer of prior computations, effectively amortising both the construction and solution costs of models for novel situations \citep{gershman_amortized_2014, dasgupta_memory_2021}.%,huys2015interplay}. 

\begin{figure}[ht]
    \begin{center}
        \includegraphics[width=0.9\textwidth]{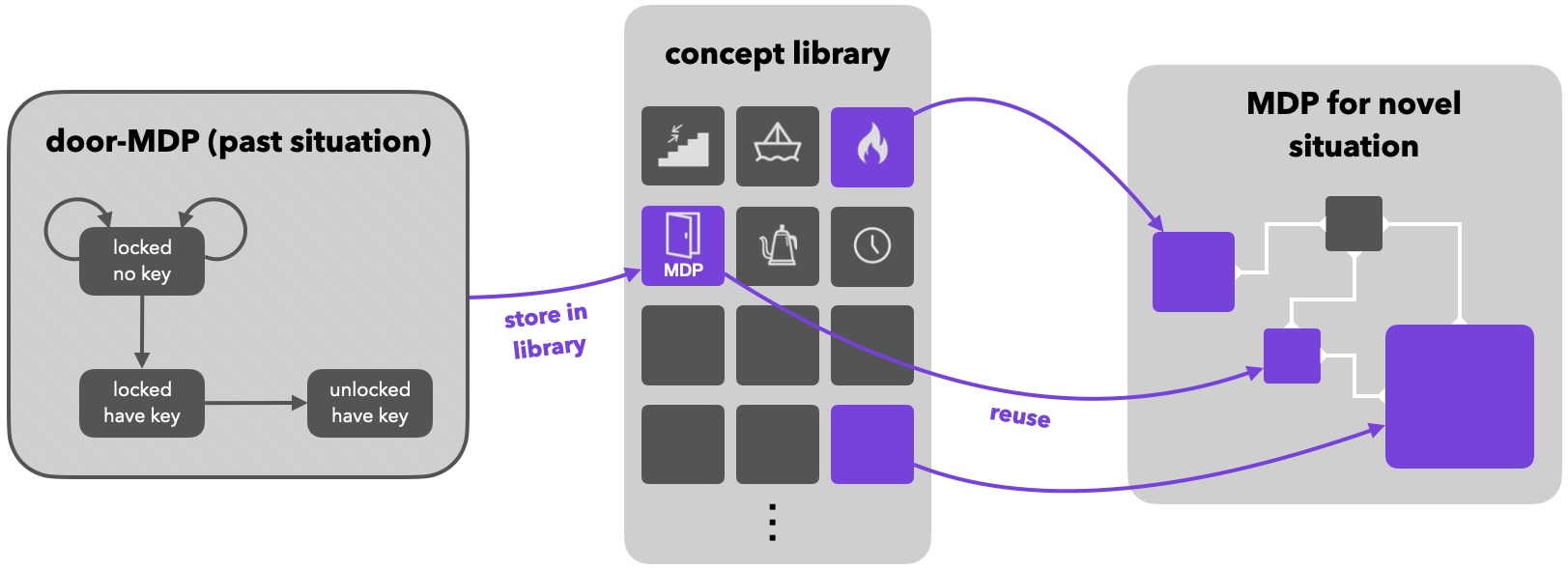}
    \end{center}
    \caption{Library of abstract MDP modules for amortised model construal.}
    \label{fig:modular-mdp}
    %\vspace{-2em}
\end{figure}

As building blocks of analogies, we propose that the brain extracts structural regularities across diverse situations in the form of reusable fragments of MDPs, and incrementally compiles them into a library of consistently useful \textit{modules}. As fragments are adapted and reused across increasingly varied contexts, they become progressively abstracted, gaining broader applicability as sources of analogy. Over time, such modules may become decoupled from their original contexts entirely, forming widely reusable conceptual primitives such as 'door', 'stairs', 'fire' or 'clock' (Fig. \ref{fig:modular-mdp}). 
Consider a small child observing that the front door (previously thought part of the wall) can be unlocked and opened by inserting and turning a specific object --- the \textit{key}. This transforms it into a functional door, similar to those inside the house. Crucially, only a specific key will work; other objects, even if similarly shaped, will fail. %Later, the child learns that opening the door of the car follows a similar process, but requires a different object, a \textit{‘car-key’}. 
Years later, the child is given their first email account. A parent explains, through an explicit analogy, that the password is the 'key' for their account --- an \textit{'email-key'}. Of course, the analogy is imperfect, e.g. there is no ‘turning’ the password, only clicking on the login button. Yet, the analogy likely helps the child greatly. For instance, they now understand that no other password, even of similar length or characters is likely to open the account. They know to guard the password, because if someone steals it they will gain access to their messages. But if they do want to grant someone access, they can simply share the password.

Despite surface differences in terms of low-level actions and states, the key/door and the email/password situations share high-level structure. By seeing the login screen as a 'door' and using the password as a 'key', the child can reuse a familiar mental representation, including transition dynamics, possible actions, and useful policies. The analogy preserves the situation’s essence, allowing the child to transfer knowledge across domains. 

In the following, we formulate the challenge of on-demand model construction, proposing analogy as a mechanism for amortising both the construction and solution costs of MDPs. We argue that humans construct models not only by analogy to single past situations, but also by recomposing abstract modules drawn from an internal library that they concurrently grow. We formalise analogies as partial MDP homomorphisms that preserve structure relevant for planning and decision making. We end with an overview of key computational challenges inherent in forming, maintaining and using such a library of composable modules.

\section{The challenge of on-demand model construction}

Unlike typical artificial RL agents, humans must not only use, but also construct their internal models of the environment. To build our formalisation of this joint challenge, we contrast the vacuum-cleaning robot with the child from our earlier examples.

\begin{figure}[ht]
    \begin{center}
        \includegraphics[width=0.7\textwidth]{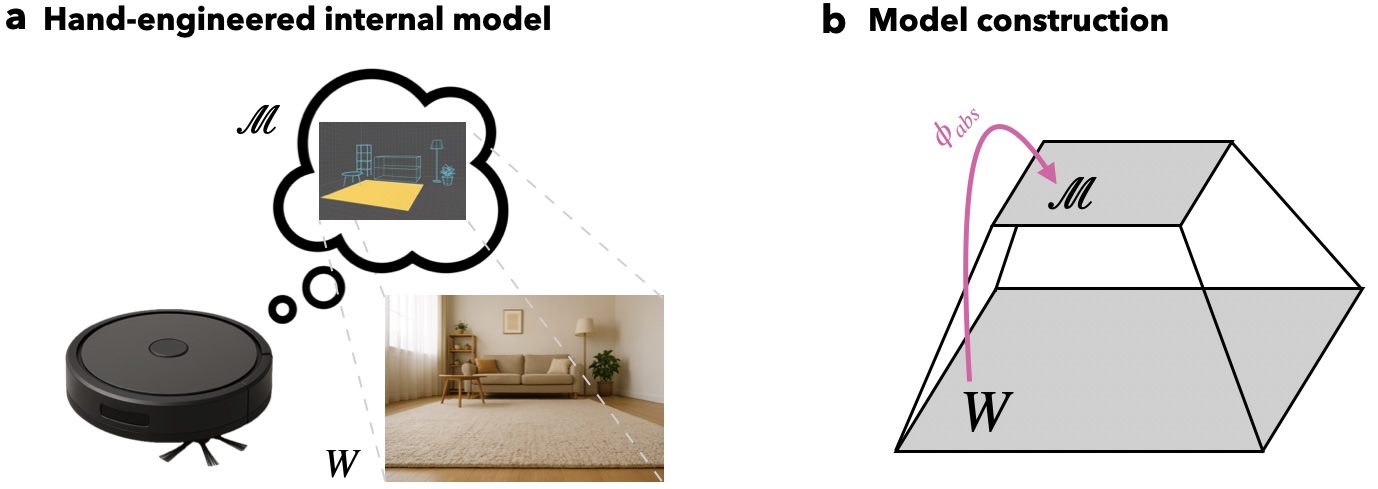}
    \end{center}
    \caption{The design process of certain real world artefacts (e.g. vacuum-cleaning robots) resembles solving the computational problem of MDP construal.}
    \label{fig:example}
    %\vspace{-1em}
\end{figure}

Consider first the perspective of a robot tasked with efficiently cleaning an apartment: it must cover all accessible floor areas, avoid bumping into furniture, and return to its charging station before the battery is depleted. In an RL setting, the subset of environmental knowledge maintained by the robot to perform this task is encoded in what we call its internal MDP $\mathcal{M}$, created by its designer. This internal MDP defines the states $s_t$ (e.g., the robot’s position, or whether a room is clean), actions $a_t$ (e.g., move forward, return to charger), the transition function $T(s_t, a_t, s_{t+\Delta t})$ (e.g., walls are impassable, the dock recharges the battery), and the reward function $R$ (e.g., for successfully cleaning a room, which could depend on the selected cleaning mode such as ‘quick clean’ or ‘deep clean’). Concretely, the robot must compute and execute a policy $\pi^*_{\mathcal{M}}$:
\begin{equation}
\pi_\mathcal{M}^*(a_t|s_t)=\arg\max_\pi \mathbb{E}[R|\pi,\mathcal{M}]
\label{eq:M_max_reward}
\end{equation}

Now let us shift perspective to the engineer responsible for designing the robot’s internal representation of the environment $\mathcal{M}$ in the first place. The first challenge for the designer is to ensure that the policy $\pi_\mathcal{M}^*$ that the robot derives using its internal MDP  survives contact with the real world. Thus, the designer must keep in mind that the optimisation in Eq.~\ref{eq:M_max_reward} is only a proxy for the true objective of maximising reward \textit{in the real world}, represented by a hypothetical ground-truth MDP $W$:
\begin{equation}
\pi_W^*=\arg\max_\pi \mathbb{E}[R'|\pi, W]
\label{eq:W_max_reward}
\end{equation}
Ideally, the internal model $\mathcal{M}$ would perfectly match the world, i.e., $\mathcal{M} = W$.
However, as Jorge Luis Borges’ famous parable illustrates, a map as detailed as the territory it represents is of little practical use \citep{borges1960rigor}. Likewise, an internal model that mirrors every detail of the environment would be computationally intractable. Therefore the second challenge for the engineer is to ensure that operating the model
is tractable under the robot’s limited computational resources. Formally\footnote{Here, we use ‘solution’ in a broad sense (e.g., it may refer to computing a plan via model-based methods, learning value estimates or policies via model-free RL, or hybrid approaches that combine both), while treating computational cost in abstract terms, which may include time and energy required for solving the model.}, the cost of computation $C_s$ required to arrive at a solution  %\footnote{We use expected reward as the objective for concreteness, but in practice, alternative criteria may be more appropriate. For instance, one might optimize for robustness, safety, worst-case guarantees, or satisficing performance, depending on the application domain and risk profile.}
has to stay below the maximum computational capacity %For simplicity, we assume a hard constraint on available computational capacity, but this constraint could be soft --- for example, allowing flexible allocation of resources.
$C_{\max}(\textrm{robot})$. Taken together, we can formalise the engineer's objective as:
\begin{equation}
\mathcal{M}^*=\arg\max_\mathcal{M} \mathbb{E}[R'|\pi^*_\mathcal{M}, W],\;\; s.t. \;C_s(\mathcal{M})\leq C_{\max}(\textrm{robot}),
\label{eq:construal_objective}
\end{equation}
where $R’$ represents the real-world goals that the engineer aims to achieve. A key difficulty is that there is no a priori bound on what aspects of the real world need to be included in the model to solve a given problem. In principle, anything could be related to anything and it is difficult to judge which details can be safely ignored. This is exactly a version of the frame problem \citep{dennett_cognitive_1990,icard_resource-rational_2015}, which has a long and storied history in philosophy and AI, and cognitive science more broadly \citep{butz2025contextualizing}.

In summary, the engineer’s objective is to construct an internal MDP for the robot that is tractable to solve, and whose resulting policy leads to the desired outcomes when executed %\footnote{As determined by the engineer’s predefined mapping of internal actions to the robot’s actuators.}
in the real world. Related ideas have been explored in the context of human planning, such as in the framework of \textit{value-guided construals} \citep{ho_people_2022}, and in the context of the frame problem \citep{icard_resource-rational_2015}. In the following, we refer to the resulting $\mathcal{M}^*$ as the \textit{construal}, although it is sometimes also referred to as a submodel \citep{icard_resource-rational_2015}, situation model \citep{zwaan1998situation}, or mental space \citep{fauconnier1994mental}. The process of building the construal is sometimes referred to as model synthesis \citep{wong_word_2023,ahmed_synthesizing_2025}.

However, an essential challenge is often ignored, namely that optimising Eq.~\ref{eq:construal_objective} itself incurs computational costs, denoted $C_c$. 
For the vacuum-cleaning robot, we might intuitively think of $C_c$ as the resources expended by the engineer in designing the model, including the engineer-hours available for the design. 
%For an effective yet affordable system, both types of resource constraints must be satisfied simultaneously: 
%\[
%\left[C_s(\mathcal{M}) \leq C_{\max}(\textrm{robot}) \right]\land \left[C_c(\mathcal{M},W) \leq C_{\max}(\textrm{engineer})\right] \] 
This additional constraint is especially important to consider if we now take the perspective of the human child, whom we described as both the designer and user of their own mental models. Unlike the robot, where the roles of model construction and model use are split, the human must manage both processes internally. 
Thus, we can define a combined objective for model construction we refer to as the \textit{on-demand objective}:
\begin{equation}
\mathcal{M}^*=\arg\max_\mathcal{M} \mathbb{E}[R''|\pi^*_\mathcal{M}, W],\;\; s.t. \;C_s(\mathcal{M})+C_c(\mathcal{M},W)\leq C_{\max}(\text{child}).
\label{eq:OD_construal_objective}
\end{equation}
That is, to act effectively in novel situations, the child must construct an internal model $\mathcal{M}$ that is not only useful but also computationally feasible to both create and to solve.

\section{Analogy making as amortised model construction} 

How do humans contend with the challenge of on-demand model construction? We argue that a key strategy is to reuse previously computed construals, by drawing analogies between past and present situations. These analogies amortise previous computations \citep{dasgupta_memory_2021, gershman_amortized_2014, huys2015interplay}, enabling new environments to be modelled in old terms, and previously computed successful policies to be mapped, at least approximately, to new situations. Central to this process is a library of abstract, composable fragments (i.e., modules) from which new construals can be flexibly assembled \citep[Fig.~\ref{fig:modular-mdp};][]{zhou_harmonizing_2024,rubino2023compositionality}. The contents of the library can be seen as offering a form of powerful inductive bias for the construal process, as well as including a complete, or at least a partial, solution of the resulting construal. In turn, the library is populated and refined by means of refactoring and organizing the construals as they are created. The computational cost of maintaining this library can be viewed as an upfront investment that reduces the effort required to construct future models.
%We suggest furthermore that basing the decomposition process in the roles that modules play in good solutions might be a basis for the definitions of the objects themselves, mirroring previous arguments from the theory of affordances (Gibson, 1979).

In the following, we sketch an account of how analogical mappings can support efficient model construction. First, building on the example of the child learning to use their email account, we examine a simplified case in which the construal for a new situation is built from a single abstract module. Then, we extend our discussion to scenarios where multiple modules must be composed and their respective solutions integrated into a coherent whole. Finally, we consider how such modules might be extracted, refined, abstracted, and maintained over time.

%We discuss how analogies can guide the construction of new models under real-world constraints, and how policies and action abstractions can be transferred to enable reuse of previously computed solutions. 
%Finally, we describe how the modular decomposition of past construals allows for more flexible and scalable reuse, and present a heuristic example that demonstrates how a library of such modules can support adaptive behaviour in a compositional navigation task.

\subsection{Making a single analogy}

\begin{figure}[h]
    \begin{center}
        \includegraphics[width=1\textwidth]{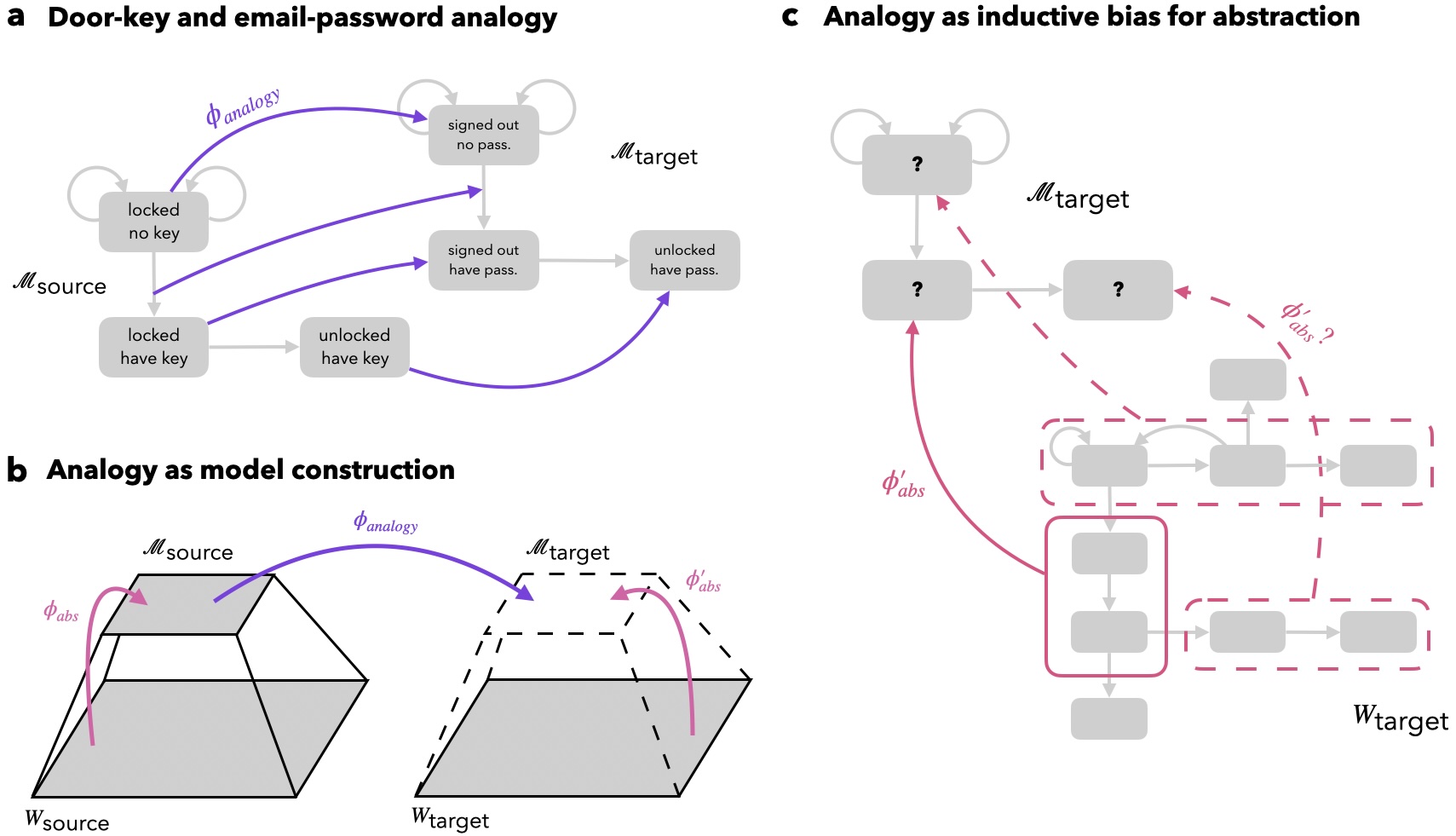}
    \end{center}
    \caption{Analogy-guided model construction. a) The child draws an analogy $\phi_{analogy}$ between the construals for the door–key ($\mathcal{M}_{\text{source}}$) and email–password ($\mathcal{M}_{\text{target}}$) situations. b) In our framework, this means constructing $\mathcal{M}_{\text{target}}$ using $\mathcal{M}_{\text{source}}$ as an inductive bias, constrained by the real-world task $W_{\text{target}}$. Here, the parent explicitly points out a previous MDP as the source. More  generally the child has to search over their library of modules (see Fig.~\ref{fig:sol-analogy}). c) In making the analogy, the child has to work out which micro-states and micro-actions correspond to which abstract states and actions. Parental guidance provides partial state–action mappings (solid pink arrows), helping the child work out the remaining ones (dashed pink arrows).}
    \label{fig:door-email}
   %\vspace{-1.5em}
\end{figure}

\textbf{Analogy as partial MDP homomorphism.} Recall our earlier example, in which the parent’s guidance helped a child draw an analogy from a familiar ‘door–key’ scenario (the source) to the new ‘email–password’ situation (the target). We formalise such analogies as structure-preserving mappings $\phi$ between a source $\mathcal{M}_{\text{source}}$ and a target construal $\mathcal{M}_{\text{target}}$. More specifically, we consider an analogy to be a \textit{partial (S)MDP homomorphism} \citep{ravindran_model_2002, ravindran_smdp_2003}, consisting of a state mapping $f(s)$ and an action mapping $g_s(a)$. The analogy might map the locked state of a door to the logged-out state of an email account ($f(s_{\textrm{locked, no key}}) = \tilde{s}_{\textrm{logged out, no pwd}}$) and the act of picking up the key to recalling a password ($g_s(a_{\textrm{get key}}) = \tilde{a}_{\textrm{recall pwd}}$; Fig.~\ref{fig:door-email}a). The mapping qualifies as a strict homomorphism if it preserves the reward and transition structures of the source MDP. %on a relevant subset of the state-action space. 
%\begin{align}
%T_{\text{target}}(f(s), g_s(a), f(s')) &= \sum_{s'' \in [s']_f} T_{\text{source}}(s, a, s''),  \label{eq:transh}\\
%R'(f(s), g_s(a)) &= R(s, a).
%\label{eq:homomorphism}
%\end{align}
%Here, $f(s)$ maps a source state $s$ to its abstract counterpart in the target MDP, and $g_s(a)$ maps the source action $a$, possibly in a state-dependent way, to an abstract action. 
%Here, $[s']_f$ denotes the set of all source states that are mapped to the same state $s'$ under $f$. Thus, Eq.~\ref{eq:transh} ensures that the transition structure $T'$ is consistent with that of the source: the probability of transitioning to the state $f(s')$ in the target is the sum of probabilities of transitioning to any of the source states in $[s']_f$. 
%Technically, this generalises the classical homomorphism condition from deterministic systems to the stochastic setting.
Intuitively, this can be seen as a generalisation of the deterministic homomorphism conditions to the stochastic case, where the homomorphism must commute with the system dynamics: applying the dynamics and then the homomorphism yields the same result as applying the homomorphism first and then the abstract dynamics \citep{hartmanis1966algebraic}. 
%The second equation ensures that the expected reward for the target state-action pair matches that of its counterpart.
Crucially, analogies typically involve partial mappings \citep{gentner_structure-mapping_1983, gentner_analogy_2017, lakoff_metaphors_2008}, preserving only aspects of structure deemed relevant in the current context. We refer to $\phi = (f(s), g_s(a))$ as a \textit{partial} MDP homomorphism when the conditions are only required to hold over a relevant subset of the state-action space rather than globally over the entire MDP.

\textbf{Construal by analogy.} From the perspective of on-demand model construction, making an analogy involves the child constructing a new internal model $\mathcal{M}_{\text{target}}$ by using $\mathcal{M}_{\text{source}}$ as an inductive bias and the real-world task $W_{\text{target}}$ as a constraint (Fig. \ref{fig:door-email}b). For simplicity, we assume that the child already knows $W_{\text{target}}$ (i.e., has access to the relevant low-level dynamics) but lacks effective abstractions to cope with its complexity. In more realistic scenarios where the low-level MDP must also be learned, the abstract structure provided by analogies could additionally guide exploration. 
Constructing the new MDP can be seen as attempting to reuse the high-level structure of the source $\mathcal{M}_{\text{source}}$ for the target construal of $W_{\text{target}}$, but with redefined state and action abstraction functions. Thus, in this simplified setting when the analogy source is explicitly provided, the central computational challenges are to work out the entirety of $\phi_{\text{analogy}}$ mapping the abstract models, and the new $\phi'_\text{abs}$ mapping low level target states and actions to the abstract model's macro-states and macro-actions (Fig.~\ref{fig:door-email}c). When a parent says, 'the password is like a key', we hypothesise that they are effectively supplying part of these mappings, implying for example that 'unlocking' has to do with typing the password on the keyboard. %specifically, a correspondence between a subset of the states and actions. %CW: this could be related to Gergely & Csibra's natural pedagogy, where parents tell children "an airplane has wings", to indicate essential mappings, rather than saying an "airplane has windows", which provide non-essential mappings
This partial mapping lowers the computational burden for the child, making it easier to infer the rest (Fig.~\ref{fig:door-email}c). In cases without external instruction, the child might find new analogies by searching over their pre-existing library of abstract modules and considering candidates according to the above process, which may also be amortised \citep{ellis_dreamcoder_2023, odonnell_fragment_2009}.

\textbf{Mapping solutions through analogies.} A key benefit of construal through analogy is that it enables amortisation of the planning process \citep{huys2015interplay}, %CW: adding this here, since we cut it elsewhere
notably when action abstractions from the source module are also effective for planning in the new construal. More generally, we propose that modules in the library encode not just MDP fragments, but also associated partial solutions (e.g., policies or policy fragments) that can also be transferred via analogy. For example, interpreting a login screen as a locked door immediately constrains a child’s policy search: actions like clicking ‘login’ without a password are expected to be futile, and many unproductive strategies can be ruled out. Instead, the child's search can be biased toward solutions involving password acquisition, providing an inductive constraint that narrows the solution space even when the exact details are different.

Importantly, generalisation to new situations often arises through much more mundane analogies. Interpreting the door of a previously unknown building---or even a novel object like a spaceship---as an instance of the abstract 'door' module immediately suggests a range of reasonable actions: to unlock, open, close, lock, go through, change the lock, or even bang on it, but not, for example, to walk through it while it is closed, attempt to unlock it with an arbitrary key, or try to plead with it. Drawing an analogy through a module to represent a new object (e.g., identifying it as a new instance of a 'door') is closely related to perceiving its \textit{affordances} \citep{gibson_theory_2014}. In Gibson’s sense, the environment is perceived in terms of the possibilities for action it offers to the agent. An affordance, like an action abstraction associated with a module, reflects not only properties of the object (or environment), but also the capabilities of the agent: for example, a lake is 'support' that affords standing-on for a water bug but not for a human, and a tall bench might be a 'chair' that affords sitting for an adult but not for a small child. From this perspective, the cognitive mechanism that maps parts of the environment to MDP modules can also be understood as a mechanism for affordance perception.
 
%We suggest furthermore that basing the decomposition process in the roles that modules play in good solutions might be a basis for the definitions of the objects themselves, mirroring previous arguments from the theory of affordances (Gibson, 1979).

\citet{ravindran_model_2002} show that, in straightforward cases where part of the abstract structure is shared exactly, if $\pi_{\text{source}}(s,a)$ is an optimal policy for the source MDP, then the mapped policy $\pi_{\text{target}}(f(s), g_s(a))$ is also optimal for the target. %, provided the homomorphism conditions hold exactly over the subset of state-action pairs that are retained by the analogy. 
Losses in performance arising from only approximate homomorphisms can be quantified via Bounded-parameter MDPs \citep{ravindran_model_2002}.

\subsection{Model construction through composition of modules}

Thus far, we have considered model construction via analogy with a single module. However, the diversity of real-world situations demands a more flexible strategy for reuse: the ability to decompose previously formed construals into fragments, and to recombine these fragments in novel configurations. By compiling a library of MDP modules $\mathcal{L}=\{\mu_i\}$ that preserve solution-relevant structure, it is possible to acquire an abstract vocabulary for model construction as well as the solution of the resulting models. 
Following previous work viewing amortisation via a library as an inference problem \citep{odonnell_fragment_2009,ellis_dreamcoder_2023, zhao_model_2023,bowers_top-down_2023,zhou_harmonizing_2024,rule2020child}, we decompose the derivation of the policies $\pi_{1:t}^*$, the construals $\mathcal{M}_{1:t}$ and the library $\mathcal{L}$ on the basis of $W_{1:t}$ into three separate stages that the agent alternates between (Fig.~\ref{fig:library-graphical-model}): %the posterior $P(\pi_{1:t}^*,\mathcal{M}_{1:t},\mathcal{L}|W_{1:t})$ into three separate stages that the agent alternates between (Fig.~\ref{fig:library-graphical-model}):
\begin{enumerate}
    %\item $P(\mathcal{M}_t|\mathcal{L},W_t)$ --- Conditional on a fixed library $\mathcal{L}$, what is the construal $\mathcal{M}_t$ for $W_t$?
    %\item $P(\pi_t^*| \mathcal{M}_t,\mathcal{L})$ --- Conditional on a fixed construal and the library, what is an effective policy?
    %\item $P(\mathcal{L}| \mathcal{M}_{1:t},\pi_{1:t}^*)$ --- Conditional on the history of construals and solutions, what modules should be in the library?
    \item $\mathcal{M}_t|\mathcal{L},W_t$ ---  Conditional on a fixed library $\mathcal{L}$, what is the construal $\mathcal{M}_t$ for $W_t$?
    \item $\pi_t^*| \mathcal{M}_t,\mathcal{L}$ ---  Conditional on a fixed construal and the library, what is an effective policy?
    \item $\mathcal{L}| \mathcal{M}_{1:t},\pi_{1:t}^*$ --- Conditional on the history of construals and solutions, what modules should be in the library?
\end{enumerate}

In the following, we briefly discuss key computational challenges raised by the above questions. 

\textbf{Composing modules and adapting solutions.} 
Since analogies can map between arbitrary subcomponents of an MDP, construals for novel situations may be assembled piecewise, drawing components from distinct previous situations ($\mathcal{M}_{t_i}$) or abstract modules ($\mu_i$). For example, a child familiar with their apartment may interpret an office door by analogy with their apartment door, while drawing on their experience with a school projector to understand a similar device in a conference room. This compositional reuse allows useful state and action abstractions to be imported in parts, enabling efficient planning in unfamiliar situations, echoing recent proposals that modular building blocks support zero-shot generalisation in novel environments \citep{bakermans_constructing_2025}. 

\begin{figure}[ht]
    \begin{center}
        \includegraphics[width=0.7\textwidth]{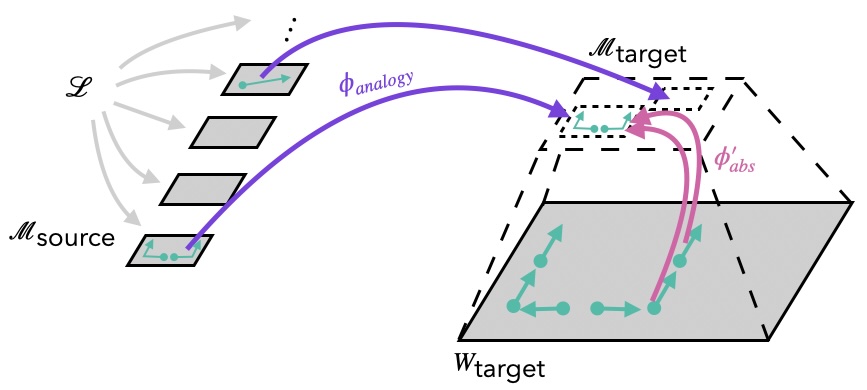}
    \end{center}
    \caption{Modular construction of model, and transfer of solutions through the analogy.}
    \label{fig:sol-analogy}
\end{figure}

A key challenge in composing modules is that while each module encodes an approximate solution to the frame problem within its original context, there is no guarantee that these solutions will transfer smoothly to new configurations. For instance, while a chair might consistently afford sitting across all of a child’s previous experiences, in a novel context---perched on a pedestal in a museum---the same actions may lead to undesirable outcomes. Therefore, the transferability of policies and action abstractions associated with a given module (e.g., the chair), or even the choice of the module from the library, can also be dependent on other modules in the construal (e.g., the pedestal). These relations between modules may be possible to represent and learn separately, as in prior approaches to compositional MDPs \citep{diuk_object-oriented_2008,guestrin_generalizing_2003, bapst_structured_2019}.

While we technically distinguish between the stages of construal and solution, these processes are tightly intertwined. A good construal can render the solution trivial, while a poor one may make a solution intractable or even impossible.  
For example, in a new situation, the optimal policy might rely entirely on a single transferred action abstraction (which involves mapping the action abstraction into the target domain's low level actions), making policy derivation straightforward and effectively reducing the problem to creating the construal. Conversely, if the imported abstractions are more granular or less applicable to the new situation, more of the computational burden is deferred to the solution phase. In either case, creating the construal is not merely a precursor to solving the problem, but often the first, and most critical step in the solution process itself.

%\textbf{Monitoring the construal.} One way to mitigate
%jumping scales, mechanism of how a tool works is ignored, until it stops working. then one needs to switch to a more fine-grained representation.

\textbf{Module extraction and refinement.}
While partial MDP homomorphisms allow reuse of components directly from past construals stored in memory, this process is made more efficient by extracting consistently useful fragments into abstract, stand-alone modules.
By keeping track of the aspects of the module that are frequently discarded in mapping to new situations, one might refine the module to serve as better source of analogies. For example, while the 'key' module might originally be extracted into a module that applies to various physical key-like objects (car key, room key, etc.), in the course of forming more distant analogies (including with passwords), the child will likely also form a highly abstract 'key' module that allows them to interpret sentences such as 'the key insight in the theory is...'. Such a movement of the source of an analogy over time from a richly represented, concrete situation to one of a set of highly abstract concepts has been aptly described as the 'career of a metaphor' by \citet{bowdle2005career}. Formally, the increasing abstraction of amortised components resembles the evolution of a newly defined primitive in fragment grammars \citep{odonnell_fragment_2009} and in approaches to program induction \citep{bowers_top-down_2023}.  In a neuroscience context, it might be understood as the consolidation of memory traces from episodic to semantic memory \citep{kali2004off,lengyel2007hippocampal, nagyAdaptiveCompressionUnifying2025}. 

\section{Discussion}
We argued that on-demand model construction poses a uniquely demanding computational challenge. While humans appear to solve this challenge with remarkable flexibility, our account suggests that this flexibility is grounded not in general-purpose reasoning, but in powerful inductive biases derived from past experience, encoded as reusable modules and structured analogies. The severity of the on-demand model construction challenge—compounded by the frame problem—likely necessitates amortisation of computational costs not only within individuals, but across individuals, societies, and generations \citep{wu2022representational}. We propose that analogies enable especially useful and abstract building blocks of construals and policies to become embedded in cultural products—stories, idioms, images, and physical artifacts \citep{velez2022representational}. This allows the creation of a cultural repository of reusable abstractions, and consequently ways of seeing situations and solving problems \citep{wu2024group}.

Despite recent advances in learning world models for RL agents \citep{schrittwieser2020mastering,hafner2025mastering}, current methods still struggle with flexible adaptation to novel situations. Approaches focused narrowly on task-relevant representations often fail to generalise even across minor reconfigurations of the same environment, though reconstruction or self-supervised objectives can offer some improvement \citep{anand_procedural_2021, hafner2025mastering}. However, even modest changes in goals, object identity or layout can still cause severe drops in performance. In contrast, theory-based RL (TBRL) methods can sometimes support the kind of structured generalisation observed in humans \citep{tsividis2021human, pouncy_inductive_2022, pouncy_what_2021}, but typically aim to recover a relatively complete model of the environment’s dynamics rather than simplified, task-specific construals --- thereby side-stepping the frame problem, but scaling poorly in more complex domains. More recent TBRL methods have begun to construct abstract simulators in limited domains by offloading parts of the abstraction process to large language models \citep[e.g.][]{ahmed_synthesizing_2025, wong_learning_2023, wong_word_2023}.

While we have proposed a conceptual framework for analogy-guided model construction, much remains to be done to translate this sketch into a concrete computational theory. Key challenges include formalizing the processes by which modules are extracted, composed, and abstracted over time, and identifying mechanisms for searching over analogies that are both flexible and computationally tractable. In particular, it remains an open question how agents might balance the reuse of familiar structure against the discovery of genuinely novel construals in environments where no clear analogy applies.
Nevertheless, if modular reuse through analogy underlies the human ability to construct useful models under resource constraints, then formalising these mechanisms may offer a path toward more robust artificial agents.

\subsubsection*{Acknowledgments}
The authors thank Mihály Bányai for helpful discussions and comments on the manuscript, and Ryutaro Uchiyama, Tianyuan Teng, Rahul Bhui, Lionel Wong, Tyler Brooke-Wilson and Joshua Tenenbaum for insightful discussions. DGN, HZ, \& CMW are supported by the German Federal Ministry of Education and Research (BMBF): Tübingen AI Center, FKZ: 01IS18039A, funded by the Deutsche Forschungsgemeinschaft (DFG, German Research Foundation) under Germany’s Excellence Strategy–EXC2064/1–390727645, and funded by the DFG under Germany's Excellence Strategy – EXC 2117 – 422037984.
%%%%%%%%%%%%%%%%%%%%%%%%%%%%%%%%%%%%%%%%%%%%%%%%%%%%%%%%%%%%%%%%
%% NOTE: THIS MARKS THE END OF THE "MAIN TEXT"
%%%%%%%%%%%%%%%%%%%%%%%%%%%%%%%%%%%%%%%%%%%%%%%%%%%%%%%%%%%%%%%%

%%%%%%%%%%%%%%%%%%%%%%%%%%%%%%%%%%%%%%%%%%%%%%%%%%%%%%%%%%%%%%%%
%% Bibliography
%%%%%%%%%%%%%%%%%%%%%%%%%%%%%%%%%%%%%%%%%%%%%%%%%%%%%%%%%%%%%%%%
\bibliography{references}
\bibliographystyle{rlj}

%%%%%%%%%%%%%%%%%%%%%%%%%%%%%%%%%%%%%%%%%%%%%%%%%%%%%%%%%%%%%%%%
% AUTHOR: If your paper has no supplementary materials, you may 
%         comment out the line below, which creates the title for
%         the supplementary materials.
%%%%%%%%%%%%%%%%%%%%%%%%%%%%%%%%%%%%%%%%%%%%%%%%%%%%%%%%%%%%%%%%
\beginSupplementaryMaterials

\renewcommand{\thefigure}{S\arabic{figure}} %CW: adding S prefix to figure/table labels
\setcounter{figure}{0}
\renewcommand{\thetable}{S\arabic{table}} 
\setcounter{table}{0}
%cw: its great we have an SI. Let's make good use if to put less than necessary details thst slow down the maintext or dont fit within the page limit

%\section{Notation}

%\begin{table}[htbp]
%    \caption{Notation}
%    \begin{center}
%        \begin{tabular}{ll}
%            \multicolumn{1}{l}{\bf Symbol}  &\multicolumn{1}{l}{\bf Definition}
%            \\ \hline \\
%            $W$         & ground truth model of the environment, formalised as an MDP (World or Worker) \\
%            $\mathcal{M}$  & the agent's construal of the current environment, formalised as an MDP \\
%             $\pi^*_\mathcal{M}$  & optimal policy for $\mathcal{M}$ \\
%             
%             $C_{s}(\mathcal{M})$  & computational cost of solving MDP M \\
%             $C_{c}(\mathcal{M})$  & computational cost of creating the construal M \\
%
%             $f(s) $  & mapping from base to target MDP's states \\
%             
%             $g_s(a) $  & mapping from base to target MDP's actions \\
%             
%             $\phi((s,a))=(f(s),g_s(a)) $  & MDP homomorphism (mapping between base and target MDP states and actions) \\
%
%             $\Phi = \langle \mathcal{M}_{base}, \mathcal{M}_{\text{target}}, \phi \rangle $  & analogy \\
%        \end{tabular}
%    \end{center}
    %\caption{Sample table caption}
%    \label{tab:exampleTable}
%\end{table}

%\begin{figure}[ht]
%    \begin{center}
%        \includegraphics[width=0.8\textwidth]{constr-with-library.jpg}
%    \end{center}
%    \caption{Construal via library}
%    \label{fig:const-library}
%\end{figure}

\begin{figure}[ht]
    \begin{center}
        \includegraphics[width=0.7\textwidth]{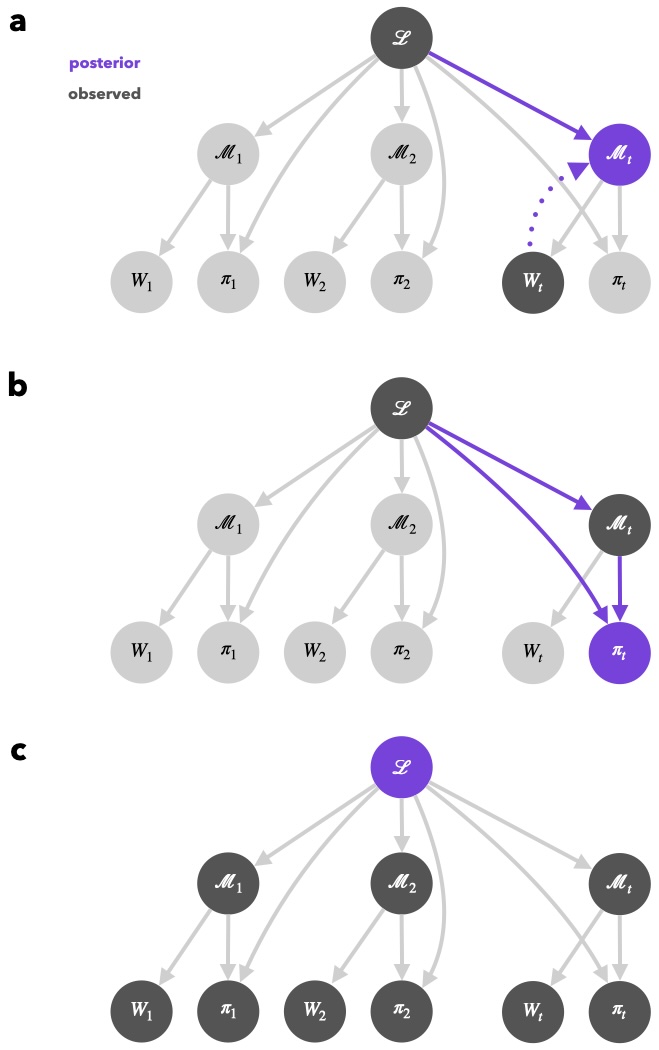}
    \end{center}
    \caption{Library building as bayesian inference.}
    \label{fig:library-graphical-model}
\end{figure}

\end{document}